\documentclass[10pt]{article}

\usepackage[utf8]{inputenc}
\usepackage[T1]{fontenc}
\usepackage{amsmath}
\usepackage{amssymb}

\usepackage[margin=2cm]{geometry}
\usepackage{amsthm}       
\usepackage{dsfont}       
\usepackage{graphicx}     
\usepackage{stmaryrd}     
\usepackage{color}
\usepackage{subeqnarray}
\usepackage{subfigure}

\definecolor{test}{rgb}{0.64,0.16,0.16}
\definecolor{vertsombre}{rgb}{0.00,0.57,0.1}
\definecolor{freeblue}{rgb}{0.25,0.41,0.88}

\newcommand{\nc}{\newcommand}
\nc{\argmin}{\mathop{\mathrm{arg\,min}}}

\nc{\ind}{\mathds{1}}

\theoremstyle{remark}
\theoremstyle{definition}

\title{Subsumption-driven clause learning with DPLL+restarts}
\author{Olivier Bailleux \\ LIB, Université de Bourgogne \\ \texttt{olivier.bailleux@u-bourgogne.fr}}
\date{Research report, June, 2019}

\begin{document}

\setlength{\parskip}{0.2cm}

\maketitle

\begin{abstract}
We propose to use a DPLL+restart to solve SAT instances by successive simplifications based on the production of clauses that subsume the initial clauses. We show that this approach allows the refutation of pebbling formulae in polynomial time and linear space, as effectively as with a CDCL solver.
\end{abstract}

\section{Introduction}

Complete SAT solvers make deductions until they find a model or produce the empty clause. 
In DPLL and CDCL solvers, these deductions are produced using assumptions generally called decisions. In DPLL solvers \cite{Davis:1962:MPT:368273.368557}, the knowledge accumulated since the beginning of the search is represented by the \emph{phases} of decision literals. Each new conflict induced by decisions increases the amount of information being accumulated.
This amount of information can be interpreted as a proportion of search space already explored that is known not to contain a model. But in the case of CDCL solvers \cite{Moskewicz:2001:CEE:378239.379017} \cite{Zhang:2001:ECD:603095.603153}, the non-chronological backtracking does not allow such a simple measurement of the amount of deduced information.
In any case, as soon as restarts are made, the information represented by the clauses produced during successive search sessions cannot be added, because the corresponding parts of the search space can overlap.

So a question arise. How to determine the relevance of the information inferred from a formula $\Sigma$, its usefulness for producing a refutation or finding a model  of this formula?
In this study, we focus on a particular form of "useful" information, namely any deduced clause that subsumes one of the clauses of the original formula.
We propose to guide the deductions made by a SAT solver to produce such clauses, so as to gradually simplify $\Sigma$.

We will focus specifically on inconsistent formulae with the aim to divide the refutation of such a formula $\Sigma$ into steps, each leading to the subsumption of a clause of $\Sigma$. The production of subsuming clauses is then considered as milestones to guide the deductions made by the solver to efficiently refute $\Sigma$. Some formulae, especially those of the pebbling type \cite{Kautz.Peeble.2004}, can be effectively refuted this way. This is of particular interest as these formulae illustrate the exponential gap between DPLL and CDCL \cite{pebble:2004}.

We propose to consider that there is a progression towards a refutation of the formula $\Sigma$ each time a clause of $\Sigma$ is subsumed.
The idea is to try to simplify $\Sigma$ by gradually reducing the size of some of its clauses until the empty clause is produced.
To implement such an approach, we use a DPLL SAT solver with restarts \cite{bailleux:hal-01985327} with an appropriate strategy for choosing decision literals.

\section{Subsumption based clause learning}

To be able to guide deductions towards the production of clauses subsuming some initial clauses of a formula $\Sigma$, we use a DPLL solver with restarts, in short DPLL+R.
Like CDCL solvers, a DPLL+R solver produces  sequences of literals that are either assumptions or deductions made by unit propagation. If $\Sigma$ is an inconsistent formula, each of these sequences results in the production of an empty clause.
Such sequences will be called contradictory AUP-sequences, where AUP stands for Assumption and Unit Propagation. To clarify this principle, we will use the following formula. 

\begin{equation}
\begin{array}{ccl}
\Sigma_1 & = & (\neg x \vee \neg y \vee r) \wedge (\neg x \vee y \vee r) \wedge (x \vee \neg y \vee r) \wedge (x \vee y \vee r)\\
 & \wedge & (\neg x \vee \neg z \vee r) \wedge (\neg x \vee z \vee r) \wedge (x \vee \neg z \vee r) \wedge (x \vee z \vee r)\\
 & \wedge & (\neg y \vee \neg z \vee r) \wedge (\neg y \vee z \vee r) \wedge (y \vee \neg z \vee r) \wedge (y \vee z \vee r) \\
 & \wedge & (\neg r \vee \neg s \vee \neg t) \\
\end{array}
\end{equation}

\begin{figure}[ht]
\centering
\centerline{\includegraphics[width=0.5\textwidth]{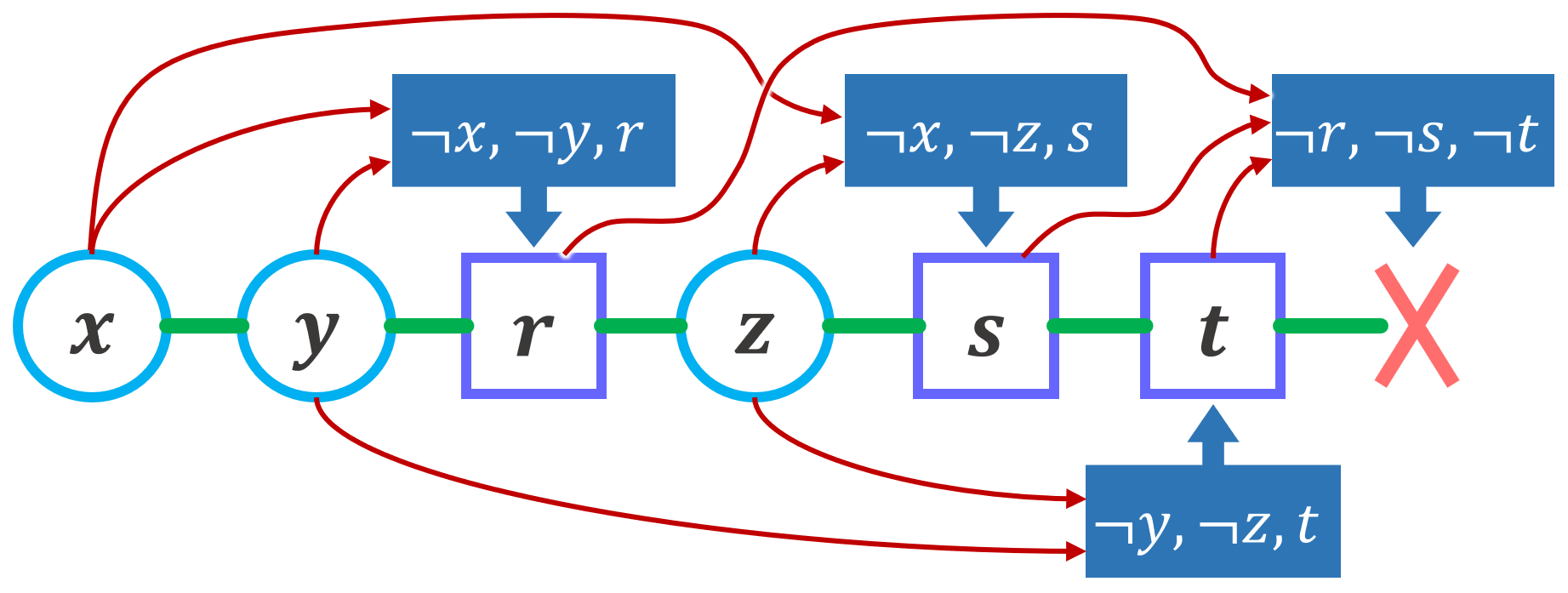}}
\caption{An example of a contradictory AUP-sequence. Circles refer to assumptions and squares to propagations.}
\label{fig:fig1}
\end{figure}

Figure \ref{fig:fig1} shows an example of contradictory AUP-sequence that could be produced by a DPLL, a DPLL+R, or even a CDCL solver from the formula $\Sigma_1$. 

DPLL+R selects its assumptions in such a way that the successive AUP-sequences of a search session constitute a tree.
A search session can be interrupted at any conflict, and the clauses representing the accumulated knowledge are then learned.

In the following, we will consider that the objective is to refute an inconsistent formula $\Sigma$, but the proof strategy we propose also allows us to find a model in the case where $\Sigma$ is coherent.

The proposed strategy is to select a clause $q$  that we will try to subsume. Let $n$ be the number of literals in $q$. A set $A$ of $n-1$ negations of literals of $q$ is used as a base of assumptions for a search session $S$.

A sequence of literals of $A$ is used to produce a AUP-sequence $T$. If $T$ is not contradictory, then it can be extended by a tree that completely explores all interpretations that have in common the assumptions of $T$. During this search, either a model is found, and in this case $\Sigma$ is consistent, or the assumptions of $T$ constitute a nogood of $\Sigma$ whose negation is a clause that subsumes $q$.

\begin{figure}[ht]
\centering
\centerline{\includegraphics[width=0.5\textwidth]{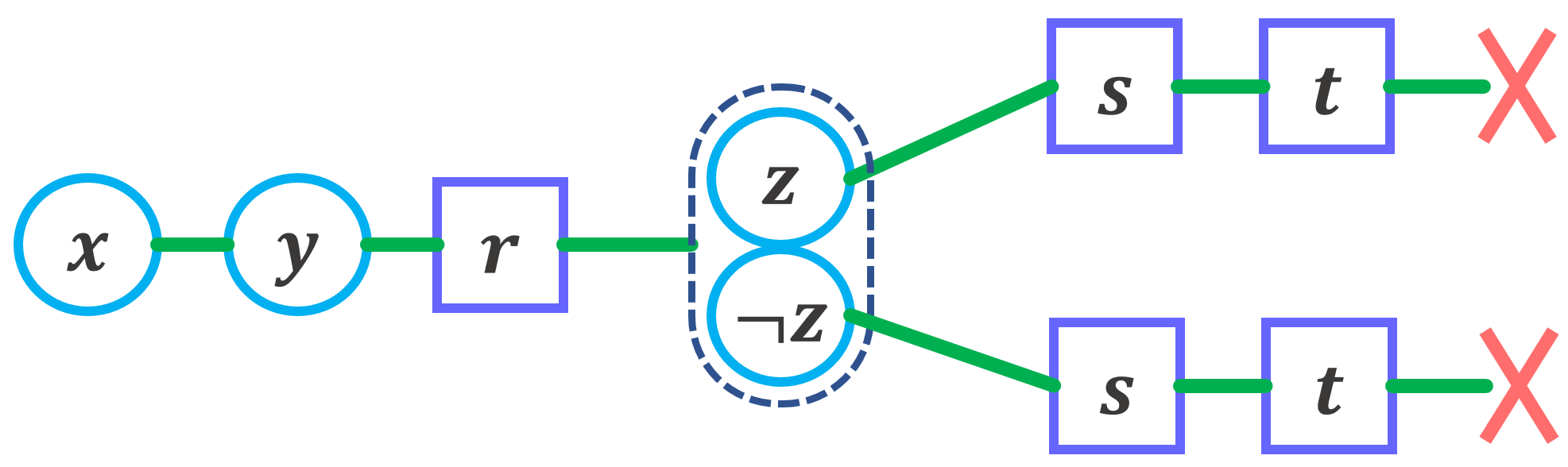}}
\caption{An example of DPLL+R session producing the clause $(\neg x \vee \neg y)$.}
\label{fig:fig2}
\end{figure}

Let us consider an example based on the formula $\Sigma_1$. Figure \ref{fig:fig2} shows a sessions of DPLL+R that produces a clause $(\neg x \vee \neg y)$ subsuming the clause $(\neg x \vee \neg y \vee r)$.

We propose to alternate phases of superficial and advanced simplifications. Superficial simplification consists in seeking clauses that can be subsumed by producing a single contradictory sequence.
Advanced simplification consists in trying to subsume a clause $q$ thanks to a partial search using as assumptions the negations of some of the literals of $q$.
This search may itself include restarts with temporary or permanent learning of new clauses. On this basis, a very large number of variants can be considered depending on the answers given to the following questions.

\begin{enumerate}
\item Should we try to simplify superficially all clauses or only some clauses, and how to choose which ones to test?
\item How to choose the clauses to be used for advanced simplification?
\item How to determine whether an attempt of advanced simplification should be completed or when it should be abandoned?
\item What restart strategy should be applied when attempting advanced simplification?
\item How to determine the literals used as assumptions during superficial and advanced simplifications? Should we try all possible arrangements?
\item How to determine when to stop the advanced simplifications in order to try if new superficial simplifications are possible?
\item How to choose the assumptions during the search phase of the advanced simplification?
\end{enumerate}

There is therefore a very large field of experimentation for the development of solvers based on this principle, which we propose to call SDCL solvers, for Subsumption Driven Clause Learning solvers.

\section{Subsumption-driven clause learning on pebbling formulae}

The inconsistent pebbling CNF formulae are particularly interesting because they illustrate the notion of exponential gap between the size of the minimum refutations that can be produced by DPLL solvers and those that can be produced by CDCL solvers. We will show that pebble formulae are solved in polynomial time by DPLL+R using only superficial simplifications.

We will only talk about or-type pebbling formulae with arbitrary arity $k$, although the properties we will prove also apply to xor-type pebbling formulae (See \cite{Nordstrom_2013} for a descriptions of such formulae).

\begin{figure}[ht]
\centering
\centerline{\includegraphics[width=0.6\textwidth]{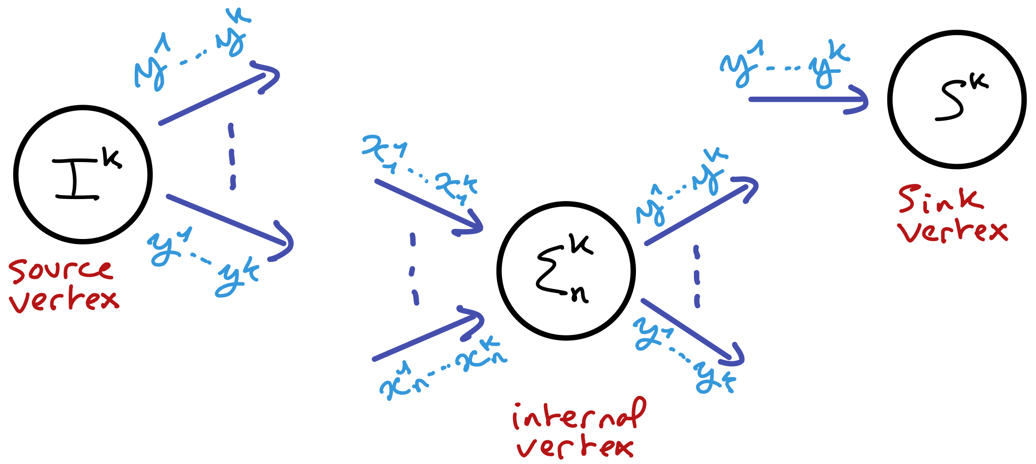}}
\caption{The three kinds of vertices of a pebbling graph related to a or-type pebbling formula with arity $k$.}
\label{fig:fig3}
\end{figure}

Essentially, a or-type pebbling formula with arity $k$ is based on an oriented acyclic graph built from vertices such as the ones presented in figure \ref{fig:fig3}. A list of $k$ variables is associated with each arc in the graph. Three types of vertices are used, namely \textit{source vertices}, \textit{internal vertices}, and \textit{sink vertices} :

\begin{itemize}
\item 
Any source vertex $I^k$ can be connected to one or more internal nodes. All the out-arcs of such a vertex are associated with the same list of $k$ variables.
\item
Without loss of generality, we will only consider graphs with a single sink vertex $S^k$, which is connected to a single internal vertex.
\item
Any internal vertex $\Sigma_n^k$ has $n$ in-arcs and one or more out-arcs. Each in-arc is associated with a different list of $k$ variables, while each out-arc is associated with the same list of $k$ variables.  
\end{itemize}
 
Each node is associated with a CNF formula. 

\begin{itemize}
\item 
Any source vertex $I^k$ with out-variables $y^1,...,y^k$ is associated with de formula $I^k(y^1,...,y^k)=(y^1\vee ... \vee y^k)$.
\item
The sink vertex $S^k$ with in-variable $y^1,...,y^k$ is associated with de formula $S^k(y^1,...,y^k)=(\neg y^1)\wedge ... \wedge (\neg y^k)$.
\item
Any internal vertex $\Sigma_n^k$ with in-variables $x_1^1,...,x_1^k,...,x_n^1,...,x_n^k$ and out-variables $y^1,...,y^k$ is associated with the formula
$\Sigma_n^k(x_1^1,...,x_1^k,...,x_n^1,...,x_n^k,y^1,...,y^k)$ that verifies the property:
\begin{equation}
\Sigma_n^k \wedge \big( \wedge_{i=1}^{n} \vee_{j=1}^{k} x_i^j \big) \models \vee_{j=1}^{k} y^j  
\end{equation}
This formula can be defined by induction as follows:
\begin{equation}
\begin{array}{lcl}
\Sigma_0^k     & = & \bigvee_{j=1}^{k} y^k \\
\\
\Sigma_{n>0}^k & = & \bigwedge_{j=1}^{k} \bigwedge_{q \in \Sigma_{n-1}^{k}} (\neg x_n^j \vee q) \\
\end{array}
\end{equation}

\end{itemize}

\begin{figure}[ht]
\centering
\centerline{\includegraphics[width=0.6\textwidth]{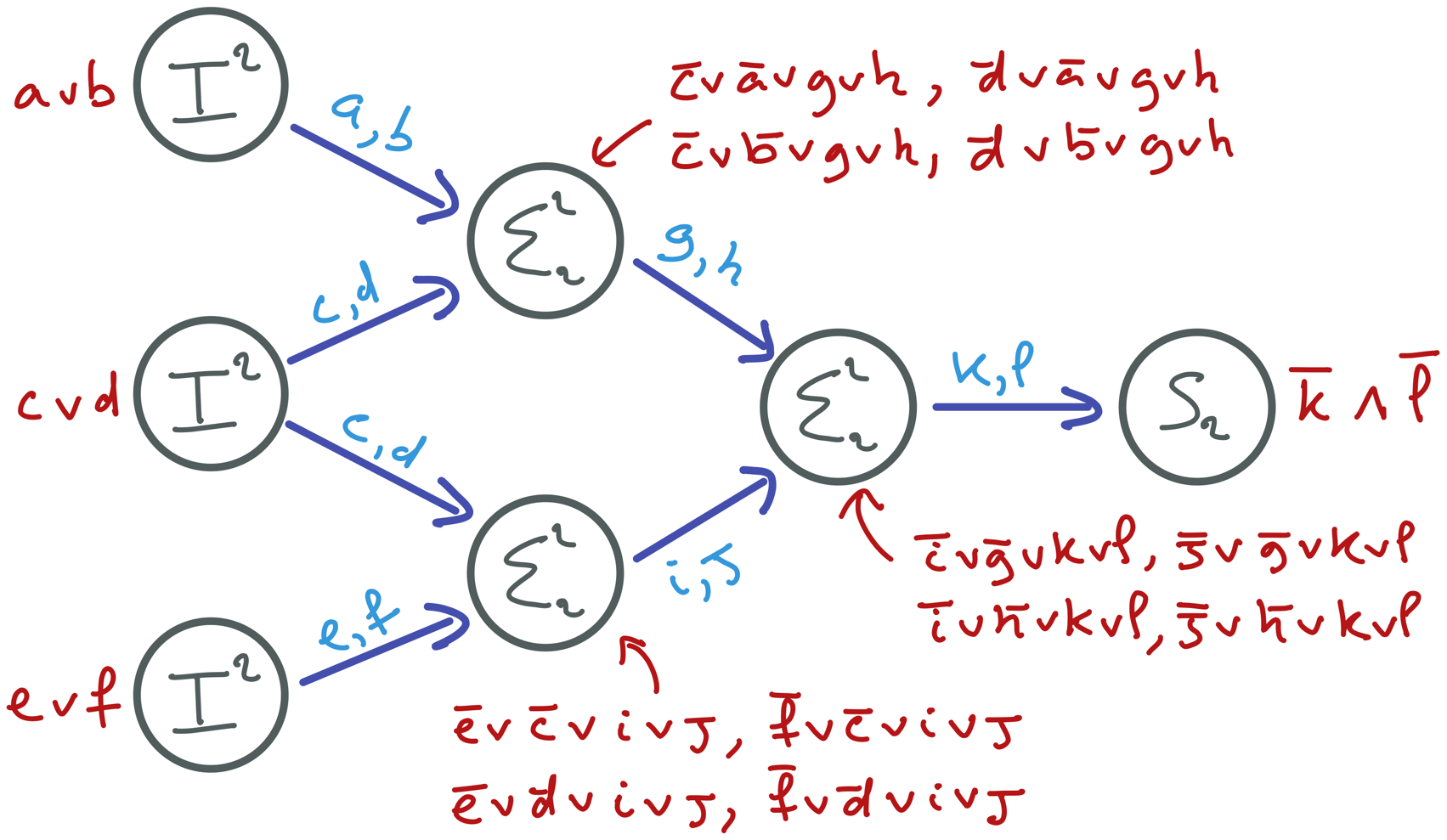}}
\caption{A pyramidal pebbling graph related to a or-type pebbling formula with arity 2.}
\label{fig:fig4}
\end{figure}

As an example, figure \ref{fig:fig4} presents the graph related to a pyramidal pebbling formula.

Now, we will show that any or-type pebbling formula $\Omega$ can be refuted thanks to a number of superficial simplifications linearly related to the number literals of $\Omega$. Each stage of the refutation aims to produce a clause $(y^1 \vee ... \vee y^k)$ from the formula $\Sigma_n^k(x_1^1,...,x_1^k,...,x_n^1,...,x_n^k,y^1,...,y^k)$ and the clauses $(x_1^1 \vee ... \vee x_1^k)$ ... $(x_n^1 \vee ... \vee x_n^k)$. Theses clauses are already available at the sources vertices. By dealing with the internal vertices in the appropriate order, they will be propagated to the sink vertex where they will conflict with the unit clauses of this vertex.

The refutation stage related to the formula $\Sigma_{n>0}^k = \bigwedge_{j=1}^{k} \bigwedge_{q \in \Sigma_{n-1}^{k}} (\neg x_n^j \vee q)$ consists in $n$ sub-stages $\Delta_n, ..., \Delta_1$, each of them producing the formula $\Sigma_{p-1}^k$ from the formula $\Sigma_{p}^k$, $p>0$, until the clause $\Sigma_0^k = (y^1 \vee ... \vee y^k)$ is produced.

Each sub-stage $\Delta_p$ consists in $\vert \Sigma_{p-1}^k \vert$ steps $\Delta_p^q, q \in \Sigma_{p-1}^k$. Each step $\Delta_p^q$ consists in a AUP-sequence with assumptions $\neg q$, which propagates $\neg x_p^1, ..., \neg x_p^k$, causing a conflict with the clause $(x_p^1 \vee ... \vee  x_p^k)$. The clause $q$, which subsumes $k$ clauses of $\Sigma_p^k$, is produced. Therefore, the sub-stage $\Delta_p$ reduces $\Sigma_{p}^k$ to $\Sigma_{p-1}^k$.

\begin{figure}[h]
\centering
\centerline{\includegraphics[width=0.6\textwidth]{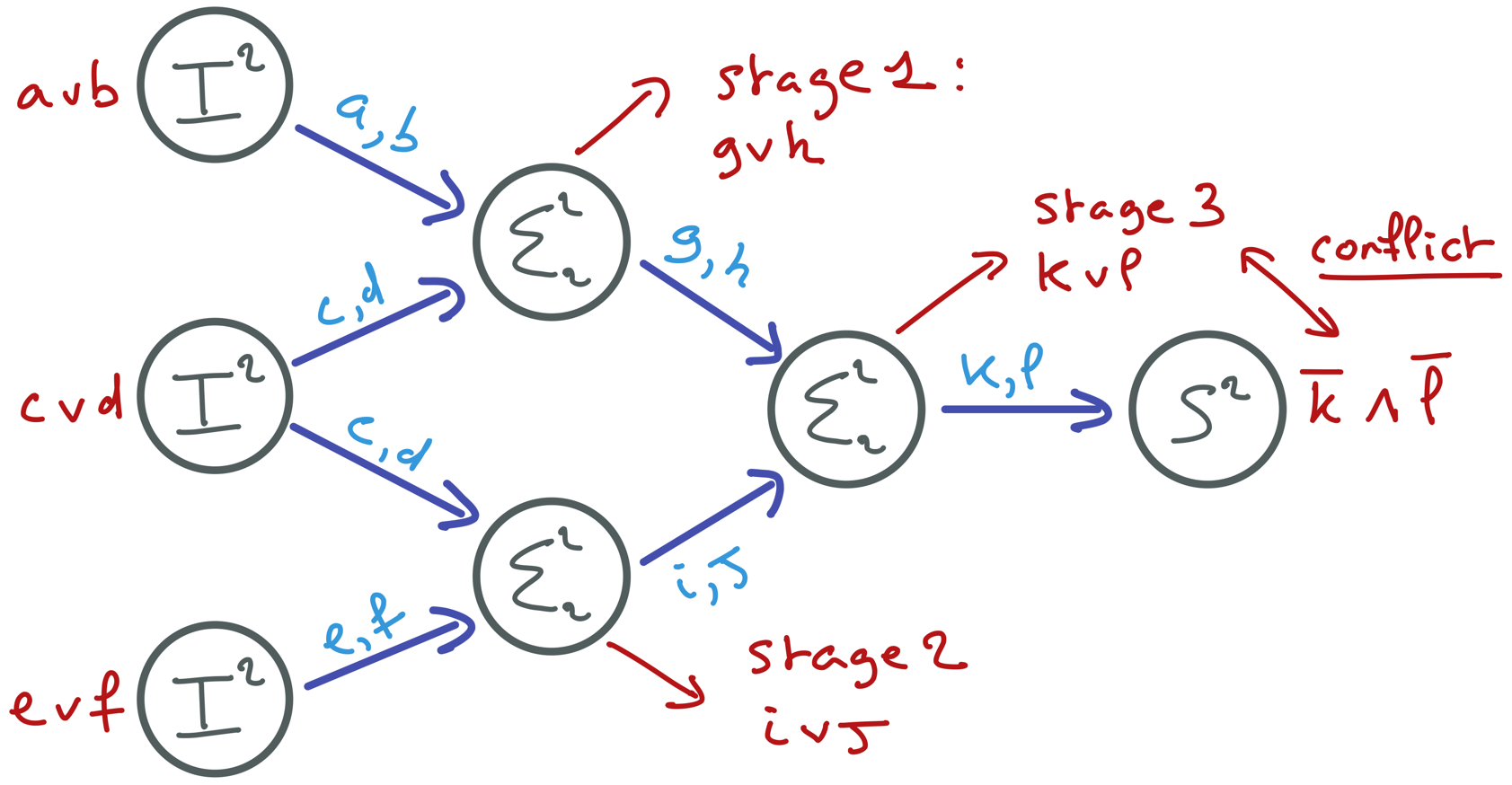}}
\caption{The 3 stages of the refutation of a pebbling formula.}
\label{fig:fig5}
\end{figure}

Let us illustrate this process on the formula described figure \ref{fig:fig4}. The three stages are shown figure \ref{fig:fig5}.

\begin{description}

  \item[stage 1]
  $\Delta_2:$ $(g \vee h \vee \neg c)$ is produced from assumptions $\neg g, \neg h, c$ and $(g \vee h \vee \neg d)$ is produced from assumption $\neg g, \neg h, d$. $\Delta_1:$ $(g \vee h)$ is produced from assumptions $\neg g, \neg h$.

  \item[stage 2]
  $\Delta_2:$ $(i \vee j \vee \neg e)$ is produced from assumptions $\neg i, \neg j, e$ and $(i \vee j \vee \neg f)$ is produced from assumption $\neg i, \neg j, f$. $\Delta_1:$ $(i \vee j)$ is produced from assumptions $\neg i, \neg j$.
  
  \item[stage 3]
  $\Delta_2:$ $(k \vee l \vee \neg i)$ is produced from assumptions $\neg k, \neg l, i$ and $(k \vee l \vee \neg j)$ is produced from assumption $\neg k, \neg l, j$. $\Delta_1:$ $(k \vee l)$ is produced from assumptions $\neg k, \neg l$.
  
\end{description}

The same principle can be applied to xor-type pebbling formulae, with the same efficiency. We do not give details, which are of purely technical interest.

\section{Synthesis and perspectives}

DPLL+restart can make the same deductions as any CDCL solver, if provided with the right assumptions. But as with CDCL, a wrong choice of assumptions can lead to the production of many useless clauses. We have proposed an assumption policy that only produces clauses subsuming existing ones, and therefore capitalizes the deduced information by simplifying the initial formula. The resulting solver has linear space complexity, like DPLL, but can solve pebbling formulae in polynomial time, like CDCL solvers.

The practical potential of this subsumption-driven learning scheme remains to be assessed on different types of SAT instances. It can only be fully informed by in-depth studies of the criteria for choosing both the clauses to be subsumed and the assumptions to be produced in order to simplify as quickly as possible the formula to be refuted. This perspective opens up a very wide, still unexplored, field of experimental and theoretical research.

\bibliography{main}
\bibliographystyle{alpha}
\newpage

\end{document}